\documentclass[10pt,twocolumn,letterpaper]{article}

\usepackage[pagenumbers]{wacv} 

\usepackage{times}
\usepackage{epsfig}
\usepackage{graphicx}
\usepackage{amsmath}
\usepackage{amssymb}
\usepackage{siunitx}

\usepackage{booktabs}
\usepackage{tabularx}

\usepackage{booktabs}
\usepackage{multirow}
\usepackage{adjustbox}
\usepackage{xcolor}
\usepackage{soul}
\usepackage{colortbl}
\usepackage{gensymb}
\usepackage{enumitem}
\usepackage{pgfplots}
\definecolor{wacvblue}{rgb}{0.21,0.49,0.74}
\usepackage[pagebackref,breaklinks,colorlinks,allcolors=wacvblue]{hyperref}

\def\wacvPaperID{2391} 
\def\confName{WACV}
\def\confYear{2027}

\title{VTF900: A 900-Subject Visible-Thermal Face Recognition Dataset and Benchmark}

\author{\parbox{\textwidth}{\centering
Jacob Abaare$^{1}$ \quad
Bianca Ianetta$^1$ \quad
Cedric Nimpa Fondje$^2$ \quad
Shuowen Hu$^3$ \quad
Benjamin S. Riggan$^{4}$
\vspace{0.1mm}
{\small
$^1$University of Nebraska-Lincoln \quad
$^2$AeroVironment \quad
$^3$DEVCOM Army Research Laboratory \quad
$^4$University of Tennessee, Knoxville}
}
}

\begin{document}
\maketitle
\thispagestyle{empty}


\begin{abstract}
   Despite advancements in visible-thermal face recognition, where thermal probe (query) face images are matched against visible enrollment templates, existing benchmarks are currently limited by small gallery sizes ($N \approx 100$ subjects).  This limitation inhibits the ability to accurately evaluate visible-thermal facial recognition at scale and to report significant recognition performance metrics, which have better operational relevance.    
   Therefore, we introduce VTF900, a large-scale visible-thermal face recognition benchmark with 900 subjects and paired imagery that includes expression, pose, and occlusion variations. 
   VTF900 is designed to support large-scale (i.e., large gallery) evaluations across verification (1:1 matching) and closed-set identification (1:N matching) for visible-thermal face recognition.  Furthermore, VTF900 specifically addresses the challenging combination of cross-spectral and cross-dataset analysis.
   In this paper, after training models on the ARL-VTF dataset, we provide benchmark analyses across convolutional (baseline), vision transformers, and domain-adaptive neural network architectures to assess how these methods scale. 
   At full gallery scale ($N = 900$), transformer models achieve strong discriminative performance under frontal conditions (\eg, ViT-B: 97.68\% AUC, 51.10\% TAR@1\%FAR), while domain-adaptive methods such as DPIT-ResNet50 demonstrate greater robustness under off-pose conditions (94.68\% AUC, 36.08\% TAR@1\%FAR). Our results show that high AUC at small scale does not translate to reliable performance at deployment scale, and that pose variation significantly amplifies performance degradation. VTF900 exposes these limitations, enabling more rigorous evaluation of cross-spectral face recognition systems.   
   
\end{abstract}

\begin{figure}
    \centering
    \includegraphics[width=0.75\linewidth]{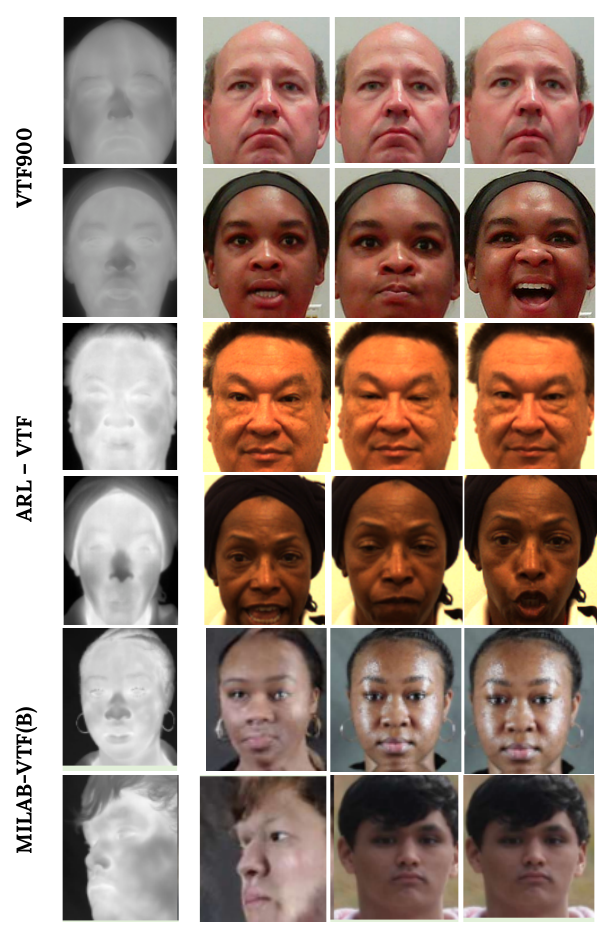}
    \caption{Sample visible-thermal pairs from VTF900 (\textbf{ours}), ARL-VTF, and MILAB-VTF(B). Each row shows one subject with the thermal probe(left) and corresponding visible images(right).  All subjects have consented to use of their images in publications.}
    \label{fig:Combine}
\end{figure}


\section{Introduction}
 
Face recognition (FR) systems have gained widespread adoption in recent years, with increasing applications across surveillance, banking, and access control~\cite{kim202650,kortli2020facesurvey}. While the performance of state-of-the-art FR using visible (0.4--0.7${\mu}m$ wavelengths) imagery has improved significantly in recent years, driven by the ubiquity of visible cameras and facial imagery~\cite{guo2016msceleb,cao2018vggface2,zhu2021webface260m}, advances in computing hardware (\eg, central processing units, graphics process units, and memory), and advances in deep learning methods~\cite{wang2021deepfr, guo2019survey, deng2019arcface}, nighttime FR, has been difficult to scale due to the limited size of evaluation datasets and galleries.

Visible-thermal face recognition (VTFR) is the process of matching thermal infrared---including midwave infrared (MWIR, 3--5$\mu$m) or longwave infrared (LWIR, 7--14$\mu$m)---probe (query) face images with visible enrollment (reference) imagery. Since the largest VTFR gallery sizes are limited to ${\sim}100$ subjects, the scalability of VTFR remains an unsolved challenge that invites further consideration due to low-light and nighttime FR benefits. However, (1) the absence of fine-grained geometric and textural patterns in thermal face images (but are present in visible face imagery) and (2) conventional perspective and occlusion differences lead to a severe domain gap between thermal and visible imagery that make VTFR inherently difficult~\cite{fondje2022learning}.  This domain gap is even more pronounced when considering VTFR with a large \emph{900-subject gallery} and a \emph{cross-dataset protocol} (\textit{i.e.}, training samples are not available from the identical data collection setup).  Therefore, the primary aim is to introduce a large-scale dataset that will enable the biometric community's advancement of VTFR at scale and under conditions that support operational feasibility assessment.  

Recent dataset contributions such as ARL-VTF~\cite{poster2021large} and MILAB-VTF~\cite{peri2021synthesis} have supported progress for VTFR by providing paired visible–thermal data for training and evaluation. 
However, these datasets do not support systematic analysis of how performance changes under large gallery sizes since both ARL-VTF and MILAB-VTF have galleries with 100 subjects or less.  These relatively smaller VTFR gallery sizes (compared to visible FR datasets) limit the ability to reliably measure the effects of gallery size and pose variation using verification (1:1 matching) and closed-set identification (forced-choice 1:N matching). 

To address the VTFR scalability issue, we introduce VTF900 which is (\textit{to the best of our knowledge}) the largest visible–thermal benchmark dataset by gallery size to date. 
VTF900 consists of 900 subjects with paired imagery including pose variations and occlusions.  
The benchmark supports evaluation across verification and closed-set identification, which enables comprehensive performance analysis including the effects of increasing gallery size. Figure~\ref{fig:Combine} provides a visual comparison between VTF900 (ours), ARL-VTF, and MILAB-VTF.

Our major contributions are: 

 \begin{itemize}[noitemsep]
    \item \textbf{VTF900}---a large-scale evaluation benchmark for cross-spectral recognition captured under diverse conditions, 
    
     \item \textbf{pose variation analysis at scale}---showing that off-pose conditions significantly degrade performance for VTFR and that domain-adaptive methods are relatively more robust to the transformer-based models baselines, 

     \item \textbf{scalability analysis}---showing the effect of increasing gallery size and how it impacts performance,

     \item \textbf{two standardized protocols}---controlled and exhaustive opportunistic protocols to mimic different operational assumptions and support reproducibility. 

 \end{itemize}

\section{Review of Literature}
This section reviews and summarizes the literature on publicly available facial recognition datasets, with particular attention to those involving thermal imagery. We describe the unique characteristics of each dataset and compare them. 
 \subsection{Multi-Modal Face Datasets}

The main challenge in cross-spectral face recognition is the domain shift--- fundamental mismatch between thermal images---caused by  differences in quality and types of patterns between imagery formed by emitted thermal radiation and reflected light. To bridge this gap, researchers have collected paired datasets using sensors that operate in the longwave infrared (LWIR, 8--14\,$\mu$m) or Midwave infrared (MWIR, 3--5\,$\mu$m) bands alongside conventional visible-spectrum cameras. Below, we review the most widely referenced and publicly available visible-thermal face datasets. We have organized them into four categories: early foundational efforts, controlled multi-condition datasets that capture varying face angles at modest scale, more recent large-scale collections that expand identity count with improved sensor resolution, and a new line of synthetic datasets generated to augment the scarcity of thermal datasets.

\subsubsection{Early and Foundational Datasets}

Early visible--thermal face datasets established the foundation for cross-spectral recognition, but were limited in both scale and variability. The \textbf{UND Collection X1/X2}~\cite{chen2003visible}, with 241 identities, was among the first to provide a meaningful subject count. However, the limited resolution (320 x 240) and number of images per subject (four) make this dataset less suitable for modern deep learning methods.

The \textbf{IRIS} dataset~\cite{abidi2023iris} builds on earlier efforts by introducing simultaneously captured visible and LWIR imagery with variation in pose and expression. However, only 30 subjects are included and limited data annotations are provided. An extended dataset, \textbf{IRIS-M3}~\cite{chang2006indoor}, increases the subject count to 88 and incorporates multi-spectral sensing, which improves robustness to illumination changes, but the overall scale remains relatively small. 

The \textbf{University of Houston (UH)} datasets~\cite{kakadiaris2005multimodal,buddharaju2007physiology} further advanced the field by combining visible, thermal, and 3D data, and later providing 7,590 MWIR images from 138 subjects. These datasets have since been widely used as benchmarks for early cross-spectral recognition research.

Similarly, the \textbf{Carl Database}~\cite{espinosa2013carl} introduced tri-spectral imaging (visible, NIR, thermal), enabling cross-domain studies, while the \textbf{SCface} dataset~\cite{grgic2011scface} focused on replicating surveillance setting conditions using low-quality sensors and varying capture distances. Despite these contributions, these early datasets remain limited in scale and image quality for training modern deep learning models.  

\subsubsection{Controlled Multi-Condition Datasets}

Subsequent datasets introduced controlled variation across pose, illumination, and expression, improving evaluation diversity but still constrained by subject count.

The \textbf{Tufts Face Dataset}~\cite{panetta2018comprehensive} provides a rich multi-modal collection ($\sim$10,000 images, 113 subjects) with strong demographic diversity. The \textbf{EURECOM} dataset~\cite{mallat2018benchmark} captures different conditions such as illumination, occlusion, and pose, but includes only 50 subjects.

Specialized datasets further explore challenging conditions. The \textbf{Sejong Face Database}~\cite{cheema2021sejong} focuses on disguise variations, while \textbf{USTC-NVIE}~\cite{wang2010nvie} targets spontaneous and posed expressions. The \textbf{RWTH-Aachen} dataset~\cite{kopaczka2018fully} provides high-resolution annotated thermal imagery but lacks paired visible data, limiting its use for cross-modal tasks. Similarly, \textbf{KTFE}~\cite{nguyen2014thermal} explores affective computing through spontaneous emotional expressions, though it is still limited to just 26 subjects. 

The \textbf{ARL-MMFD} dataset~\cite{hu2016polarimetric,zhang2019synthesis} introduces polarimetric thermal imaging, offering additional structural cues beyond intensity. The more recent \textbf{ComplexFace} dataset~\cite{he2024complexface} adds temporal variation and occlusion conditions but remains limited to 70 subjects.

Overall, while these datasets introduce richer covariates, they remain insufficient for studying performance at scale.

\subsubsection{Large-Scale and Modern Datasets}

Recent efforts aim to increase scale and better capture variability, though identity count remains a key limitation.
The \textbf{SpeakingFaces} dataset~\cite{abdrakhmanova2021speakingfaces} provides a large tri-modal collection (142 subjects) with synchronized audio and multi-view capture. The \textbf{Charlotte-ThermalFace} dataset~\cite{ashrafi2022charlotte} uniquely incorporates environmental factors such as temperature and humidity, but includes only 10 subjects. The \textbf{UL-FMTV} dataset~\cite{ghiass2018ulfmtv} is the largest MWIR video dataset with 238 subjects and long-term time-lapse variation, enabling studies of aging in the thermal domain. The \textbf{ARL-VTF} dataset~\cite{poster2021large} on the other hand represents a major step forward with roughly 500,000 paired visible--thermal images from 395 subjects, along with detailed annotations and evaluation protocols, making it a key benchmark for cross-modal verification. The \textbf{MILAB-VTF(B)} dataset~\cite{peri2021synthesis} further increases scale to 400 subjects and introduces long-range outdoor capture, incorporating atmospheric effects absent in most datasets. However, the lack of synchronization requires post-processing alignment. Despite these advances, even the largest datasets remain significantly smaller than visible-spectrum benchmarks, limiting the study of identity scaling effects.

\subsubsection{Synthetic Thermal Face Datasets}

Recent work explores synthetic data generation to address data scarcity. The \textbf{ThermVision-DB} dataset~\cite{farooq2026thermvision} uses diffusion models to generate thermal imagery, providing a scalable and privacy-compliant alternative. However, synthetic thermal signatures do not model physical radiometric properties or biometric identity, making them complementary rather than a replacement for real datasets.

\subsection{Dataset Comparison}

Across the datasets in Table \ref{tab:dataset_comparison}, a consistent pattern emerges. There is a trade-off between scale, annotation richness, and variability. While modern datasets have significantly improved in resolution and capture conditions, identity count remains the primary limitation, with most datasets containing fewer than 400 subjects for both development and evaluation.

As a result, current benchmarks do not adequately capture the challenges of large-scale cross-modal recognition. As identity count increases, impostor comparisons become more difficult, and performance degrades in critical operating regions. This highlights the need for datasets that jointly scale identity count and variability, enabling more realistic evaluation of cross-spectral face recognition systems.

\begin{table*}[tbp]
\centering
\caption{Summary comparison of publicly available visible--thermal face datasets, sorted by identity count. Synthetic datasets are marked with $^\dagger$.}
\label{tab:dataset_comparison}
\resizebox{\textwidth}{!}{%
\begin{tabular}{lcccll}
\toprule
\textbf{Dataset} & \textbf{Total (Gallery) Subjects} & \textbf{Approx. Images} & \textbf{Thermal Sensor} & \textbf{Conditions} & \textbf{Notable Features} \\
\midrule
Charlotte~\cite{ashrafi2022charlotte} & 10 & $>$10,000 & LWIR, 464$\times$348 & Pose, distance, temperature & Environmental annotations; 16-bit raw \\
KTFE~\cite{nguyen2014thermal} & 26 & -- & LWIR (NEC R300) & Expression & Spontaneous emotion; paired VT video \\
IRIS~\cite{abidi2023iris} & 30 & $\sim$4,228 pairs & LWIR, 320$\times$240 & Pose, expression & Simultaneous VIS--LWIR capture \\
Carl~\cite{espinosa2013carl} & 41 & -- & LWIR, 160$\times$120 & Pose, expression & Tri-spectral: VIS + NIR + Thermal \\
EURECOM~\cite{mallat2018benchmark} & 50 & $\sim$2,100 pairs & LWIR, 160$\times$120 & Pose, illumination, expression, occlusion & Dual-sensor simultaneous capture \\
ThermVision-DB$^\dagger$~\cite{farooq2026thermvision} & 50 & 30,200 frames & LWIR (synthetic), 512$\times$512 & Pose, expression & Diffusion-generated; GDPR-compliant \\
ComplexFace~\cite{he2024complexface} & 70 & 55,890 & LWIR, 640$\times$512 & Disguise, occlusion, time-lapse & Multi-session over 3 months \\
IRIS-M3~\cite{chang2006indoor} & 88 & -- & LWIR + Multi, 320$\times$240 & Pose, illumination, expression & Multi-spectral sensor added \\
UH (v1)~\cite{kakadiaris2005multimodal} & 88 & -- & MWIR, -- & Pose & VIS + Thermal + 3D range \\
RWTH-Aachen~\cite{kopaczka2018fully} & 94 & -- & LWIR, 1024$\times$768 & Pose, expression & High-res thermal-only; landmark + emotion \\
Sejong~\cite{cheema2021sejong} & 100 & -- & LWIR, -- & Pose, occlusion, disguise & Disguise-focused (wigs, masks, etc.) \\
ARL-MMFD~\cite{hu2016polarimetric,zhang2019synthesis} & 111 & -- & LWIR + Polarimetric LWIR & Distance & Polarimetric LWIR + VIS \\
Tufts~\cite{panetta2018comprehensive} & 113 & $\sim$10,000 & LWIR, 320$\times$240 & Pose, expression & Multi-modal (VIS, 3D, Sketch, NIR) \\
SCface~\cite{grgic2011scface} & 130 & 4,160 & NIR, surveillance & Distance, pose & Surveillance cameras at 3 distances \\
UH (v2)~\cite{buddharaju2007physiology} & 138 & 7,590 & MWIR, -- & Pose, expression & Cross-spectral recognition focus \\
SpeakingFaces~\cite{abdrakhmanova2021speakingfaces} & 142 & $>$13,000 instances & LWIR, 464$\times$348 & Pose, expression, time-lapse & Tri-modal: VIS + Thermal + Audio \\
USTC-NVIE~\cite{wang2010nvie} & 215 & -- & LWIR, 320$\times$240 & Illumination, expression, glasses & Spontaneous + posed expressions \\
UL-FMTV~\cite{ghiass2018ulfmtv} & 238 & -- & MWIR, -- & Pose, expression, time-lapse & 2--4 year time-lapse per subject \\
UND X1/X2~\cite{chen2003visible} & 241 & $\sim$960 & LWIR, 320$\times$240 & Pose & Earliest large-identity VT dataset \\
ARL-VTF~\cite{poster2021large} & 395(100) & $\sim$500,000 & LWIR, 320$\times$256 & Pose, expression, glasses & Time-synchronized; standardized protocols \\
MILAB-VTF(B)~\cite{peri2021synthesis} & 400(80) & video & MWIR, -- & Pose, distance (1.5--400\,m) & Paired VT video; long-range outdoor \\
\textbf{VTF900} $^\ast$ & 900(900) & ~240,534 & Broadband(0.7-- 5)${\mu}m$ & Pose, Expression, Mask, Glasses & High resolution \\
\bottomrule
\end{tabular}%
}

\vspace{2pt}
{\footnotesize\raggedright $^\ast$ VTF900 is intended primarily as an evaluation benchmark. Broadband facial signatures are dominated by emitting thermal radiation in the MWIR band, which is consistent with the high emissivity of human skin (0.91) ~\cite{wolff2005face,hu2015thermal}\par}
\end{table*}

\section{Collection Methodology}
This section contains details of the dataset characteristics, cameras used, and the environmental conditions under which the imaging was performed.  Also, technical specifications such as sensor details and the different capturing protocols used in the data acquisition are provided.

\subsection{Dataset Description}
VTF900 is a cross-spectral dataset of 900 subjects and ${\sim}240{,}534$
images, captured across 20 sessions under 8 pose and appearance conditions:
frontal, expression, four pose conditions (yaw-left, yaw-right, pitch-up,
pitch-down), and two occlusions (glasses, mask). Subject coverage varies by
condition: frontal, pose, expression, and occlusion conditions span 100\%, 99.4\%, 27.4\%,
and 20.2\% of subjects, respectively. Each capture is paired across two adjacent cameras and
registered to a $224\times224$ frame, averaging 115 visible and 152 broadband
images per identity. All 900 subjects are used in this paper for purposes of scalability analysis.

Data collection received Institutional Review Board approval ARL 14 - 052; all
subjects consented to research use of their data and separately indicated whether their likeness could appear in published materials. The capture protocol comprised a neutral-pose baseline followed by controlled head-movement sequences
(${\sim}10$\,s each): the subject first faced the camera with a neutral expression and all removable occlusions (glasses, masks, hats) taken off, then tilted right, left, up, and down; yaw-right, yaw-left, pitch-up, pitch-down using marked reference points to induce ${\sim}60^\circ$ rotation
for near-profile views. Subjects wearing PPE masks (occluding nose and mouth) or eyeglasses repeated the same sequence.

\begin{figure}[tbp]
    \centering
    \includegraphics[width=0.8\columnwidth]{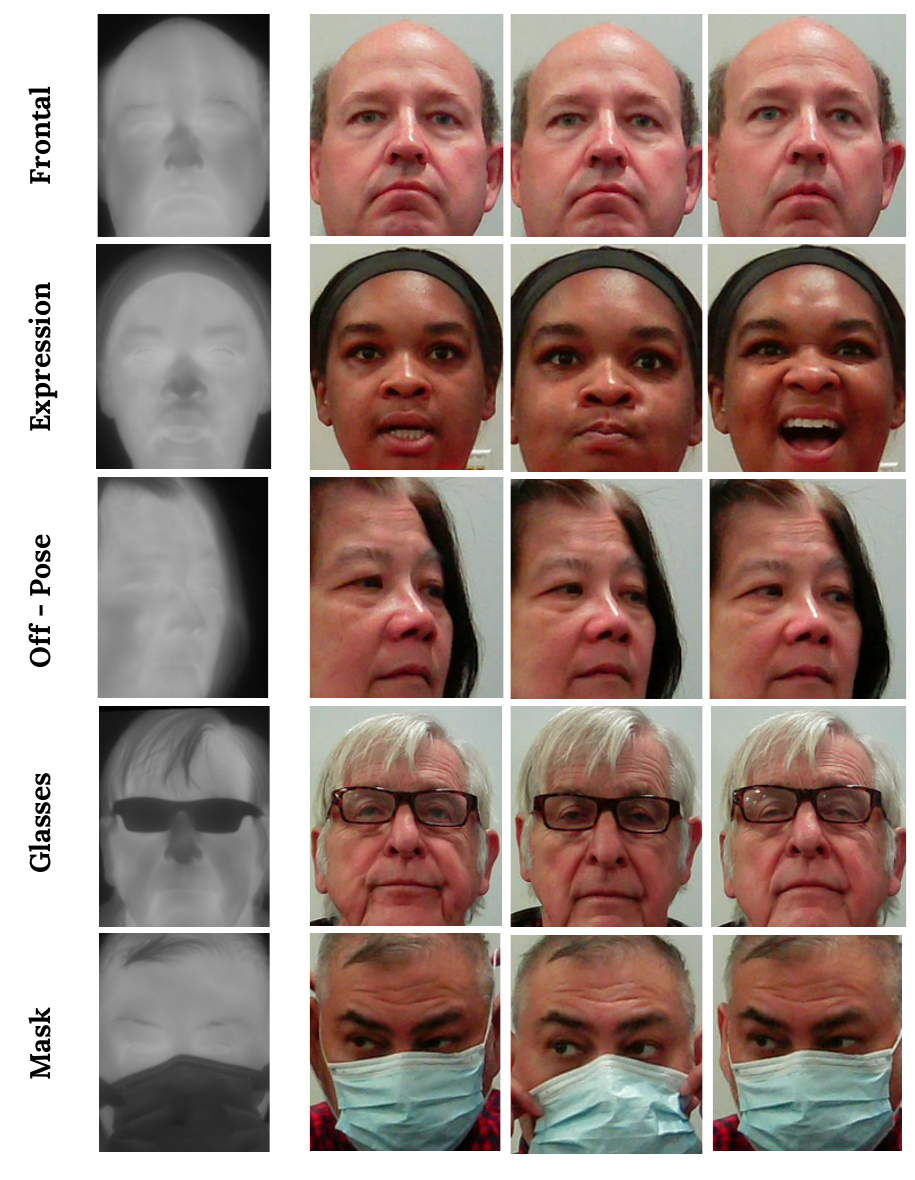}
    \caption{Overview of VTF900 showing paired visible-thermal images across different conditions. All subjects have consented to the use of their images in publications.}
    \label{fig:vtf900_overview}
\end{figure}

\textbf{Annotation:} Visible-thermal image pairs were registered using a similarity transform computed from five fiducial landmarks---the eye centers, nose base, and mouth corners. The landmarks were automatically detected using RetinaFace~\cite{deng2020retinaface} for visible imagery. However,  the canonical RetinaFace detection model does not transfer well to broadband (thermal) imagery.  Therefore, we fine-tuned RetinaFace on a large subset of raw thermal images from VTF900 and applied the resulting model to automatically detect the remaining thermal images. This yields consistent, modality-paired landmark annotations across the dataset.  After alignment, all images are cropped to $280\times360$ (width $\times$ height) pixels.  This alignment and cropping procedure is consistent with the process used for ARL-VTF~\cite{poster2021large}.
 
\subsection{Imaging Hardware and Environment}

Image acquisition was done using a dual-camera setup comprising a Canon DSLR  
visible-light camera and a Cyan CS3 broadband thermal camera, both mounted on a 
stand and positioned at eye level to center the field of view on the 
subject's face. Subjects were seated approximately 2.1 meters from the 
camera setup. The visible camera captured images at a resolution of 
$1{,}280 \times 720$ pixels.

The broadband sensor, the Cyan CS3, has resolution of 
$1{,}920 \times 1{,}080$ pixels with a 5$\mu$m pixel pitch and a 
broadband spectral response spanning 0.7--5\,$\mu$m, covering the 
near-infrared (NIR), short-wave infrared (SWIR), and mid-wave infrared 
(MWIR) bands.  Although the broadband sensor acquires some reflected energy, images acquired indoors are often dominated by the MWIR bands. The sensor achieves a noise equivalent irradiance (NEI) 
of approximately $1 \times 10^{-16}$\,W/cm$^2$, yielding high-contrast 
imagery with fine spatial detail at the acquisition distance. Additional narrowband optical filters were employed during selected capture sessions to isolate specific sub-bands within the SWIR and MWIR ranges.

All imaging was conducted in a controlled indoor environment. 
Visible-spectrum illumination was provided by artificial light sources 
mounted on adjustable stands, positioned to ensure uniform, 
high-intensity coverage of the subject's face and minimize shadowing. 
Ambient temperature was maintained at standard room temperature to 
minimize interference with thermal imaging.

\textbf{Requesting the Database:} Access to the VTF900 is granted exclusively for scientific research purposes. The dataset is available upon request.  New requests will be required to have (or obtain) a fully executed written agreement, 
such as a Cooperative Research and Development Agreement (CRADA), Cooperative Agreement (CA), or equivalent.  Additionally, a data transfer and use agreement (DTUA) will be used to outline specific data use and data security requirements for the VTF900 data. All requests are subject to review and approval based on the validity of the proposed research. Requests made by contacting Shuowen Hu at shuowen.hu.civ@army.mil.  

\section{Implementation}

To evaluate adaptability and robustness of cross-spectral face recognition models on VTF900, we consider two categories of models: a set of classical and domain-adaptive baselines and a set of transformer-based architectures~\cite{salim2026cross}.

As a baseline, we use a Naive VGG16~\cite{simonyan2014very} model without
cross-spectral adaptation to highlight the difficulty without domain alignment. We also compare this against competitive cross-spectral methods: the Residual Spectral Transform (RST)~\cite{fondje2020cross}, which applies residual transformations to intermediate feature maps to reduce the visible--thermal gap, and the Domain Pose Invariant Transform (DPIT)~\cite{fondje2022learning}, which learns representations invariant to both modality and pose. For RST and DPIT, we use the authors' publicly released weights trained on ARL-VTF; DPIT provides both VGG16 and ResNet-50 backbones. All are evaluated directly on VTF900 without further adaptation, assessing out-of-domain generalization.

We complement these with transformer models in a unified dual-stream framework: a weight-shared backbone processes both visible and thermal inputs, followed by modality-specific projection heads mapping to a 512-d $L_2$-normalized embedding.
All transformer models are fine-tuned on ARL-VTF~\cite{poster2021large} before evaluation on VTF900. We consider three backbones based on distinct design choices: ViT-B~\cite{dosovitskiy2020image} (canonical global attention), Swin-S~\cite{liu2021swin} (hierarchical, windowed attention), and DINOv2~\cite{oquab2023dinov2} (self-supervised), letting us probe how attention type and pretraining strategy affect cross-domain transfer.

To each shared embedding, we independently consider a standard linear classifier and three
margin-based losses: ArcFace~\cite{deng2019arcface},
AdaFace~\cite{kim2022adaface}, and CosFace~\cite{wang2018cosface} yielding 12 transformer variants, from which the top three by Rank-1 and AUC are reported. The three transformer models, together with the other four methods, form the seven benchmarked systems.

\textbf{Training Configuration.}
All 12 transformer variants share an identical training protocol,
leaving the backbone and head choice as the only sources of variation
across the benchmark tables. Backbones are initialized from
ImageNet-supervised (ViT-B/16, Swin-S) or self-supervised DINOv2 weights
and fine-tuned end-to-end on the ARL-VTF development split 1
($236$ identities, $224\times224$ input). We use AdamW
($\text{weight decay}=0.05$) with a base learning rate of
$5\times10^{-4}$ for the projection heads and $5\%$ of that rate for the
backbone, linear warmup followed by cosine decay, and an effective batch
size of $32$. The objective combines a per-stream classification loss
with a bidirectional cross-modal contrastive term so cross-spectral alignment is driven solely by the shared backbone and the contrastive objective. Unlike RST and DPIT, which include explicit domain adaptation and pose-invariance techniques, the transformers are adapted only through the contrastive objective.  We train for at most $60$ epochs with early stopping on validation Rank-1 (patience~$7$). All models are trained on four
RTX~2080 Ti GPUs. Full hyperparameters and augmentation are in the supplementary material
(section §A.2).

\section{Baseline Evaluations}

\label{sec:baseline}

This section presents a comprehensive baseline evaluation for the VTF900 dataset. To establish benchmark performance on VTF900, we evaluate and report the cross-domain performance on the selected models. Using two distinct protocols, we report (i) verification (1:1) and closed-set identification (1:N); (ii) probe-condition performance, including frontal vs. off-pose; and (iii) scale conditions, including how performance holds under increasing gallery size $N$ ranging from 100--900.  In every evaluation, the gallery is built from visible enrollment images and the probes are broadband (thermal).



\begin{figure*}[tbp]
    \centering
    \begin{subfigure}[b]{0.38\textwidth}
        \includegraphics[width=\linewidth]{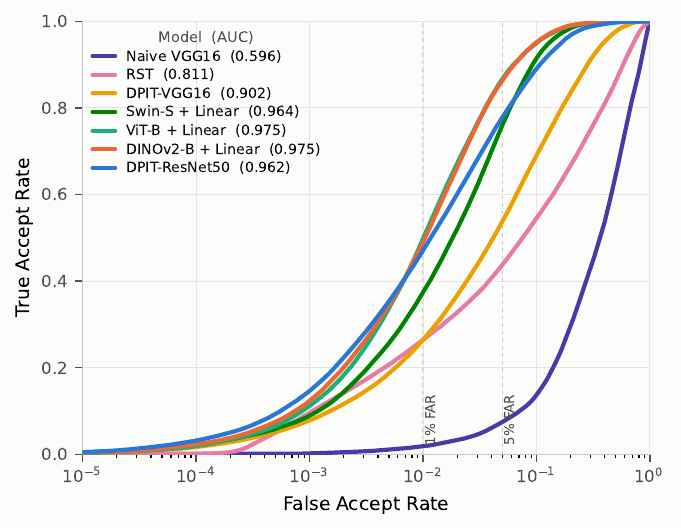}
        \caption{Frontal pose verification results.}
        \label{fig:roc_frontal}
    \end{subfigure}
    \begin{subfigure}[b]{0.38\textwidth}
        \includegraphics[width=\linewidth]{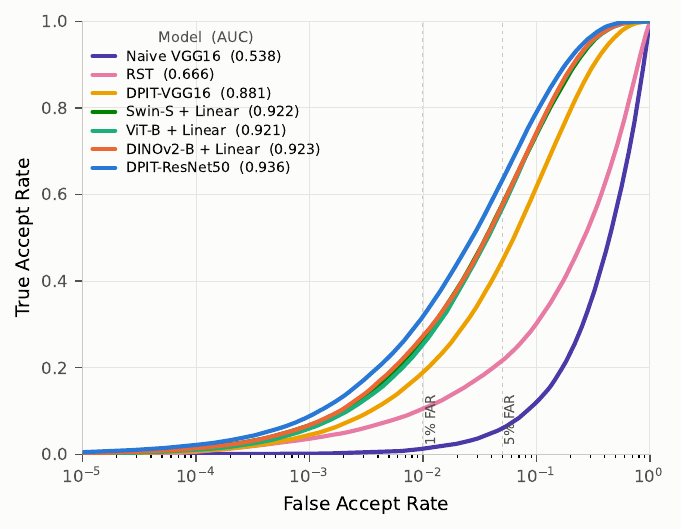}
        \caption{Off-angle pose verification results.}
        \label{fig:roc_yaw}
    \end{subfigure}
    \caption{ROC curves for cross-modal thermal-to-visible face verification under protocol A. }
    \label{fig:roc_curves}
\end{figure*}
 
\subsection{Protocols \& Gallery Construction}
We use two evaluation protocols to represent different scenarios under which the intelligence community captures images, since the processed collection yielded a different number of images for each subject.
 
\textbf{Protocol A: Controlled set  ($G_{10}$/$P_{10}$).}
A small subset of images per subject, selected evenly to span the full captured sequence, representing a standard controlled enrollment setting. Gallery enrollment (visible images) denoted as $G_{10}$, is restricted to 10 frontal images per subject across $N = 888$ \footnote{The 888 subjects are those with at least 10 frontal enrollment images, 10 frontal probe images, and 10 pose probe images; the remaining 12 are excluded for lacking at least one of those requirements.} subjects. The probe set denoted as $P_{10}$ comprises 10 frontal images (sampled disjointly from the gallery images) and 10 pose images for reproducibility. 

\textbf{Protocol B: Exhaustive Set ($G_{all}$/$P_{all}$).}
The entire set of gallery and probe images across $N = 900$ subjects, representing common surveillance scenarios from defense, intelligence, and law enforcement communities where many images from non-cooperative subjects are opportunistically acquired and enrolled. The gallery $G_{all}$ is roughly $11.7\times$ larger than in Protocol~A, with an average of $115$ visible images per subject and containing frontal, off-pose, and profile enrollment imagery. The probe (query) set, $P_{all}$ contains an average of 152 thermal images per subject. 

\textbf{Gallery construction.}
A probe $p$ is scored against identity $i$ by the maximum cosine similarity over
that identity's gallery set $\mathcal{G}_i$:
\begin{equation}
s(p, i) \;=\; \max_{g \in \mathcal{G}_i} \; \cos\!\big(\mathbf{z}_p, \mathbf{z}_g\big),
\label{eq:score}
\end{equation}
where $\mathbf{z}$ is an $\ell_2$-normalized embedding. For identification,
identities are ranked by $s(p,i)$, giving a candidate list of length $N$. $\mathcal{G}_i$ is constructed differently under each protocol.

Under \textbf{Protocol A}, each subject is represented by $K=10$ templates, one per frontal enrollment image, and every probe is scored against all of them (Eq.~\ref{eq:score}), following established cross-spectral matching practice in~\cite{riggan2016optimal, zhang2019synthesis, poster2021large}. 
For identification, we optionally reduce these to $K$ per-identity templates by $k$-means, but adopt $K=1$, as larger $K$ yields no consistent gain.

Under \textbf{Protocol B}, enrollment spans multiple perspectives (all captured conditions) rather than the frontal enrollment of protocol A. Each subject is represented by a single template ($K=1$) that averages these multi-perspective embeddings into one appearance summary. Matching and retrieval are done by probing against the template. We evaluated larger $k$-means template counts and again found
no consistent gain (supplementary material section §A.3.1), and therefore retain the averaged single-template gallery. 

\begin{table*}[tbp]
\scriptsize
\centering
\caption{Cross-modal verification under protocol A at gallery scale ($N=888$), under frontal
(F$\rightarrow$F) and pose (F$\rightarrow$Pose) probes. Best results per column
are shown in \textbf{bold}. EER is lower-is-better ($\downarrow$); all others
higher-is-better ($\uparrow$).}
\label{tab:verification_top7}
\setlength{\tabcolsep}{5pt}
\renewcommand{\arraystretch}{1.0}
\begin{adjustbox} {max width=0.8\textwidth}

\begin{tabular}{l cccc cccc}
\toprule
& \multicolumn{4}{c}{F$\rightarrow$F} & \multicolumn{4}{c}{F$\rightarrow$Pose} \\
\cmidrule(lr){2-5}\cmidrule(lr){6-9}
Model & AUC & EER & TAR@1\% & TAR@5\%
      & AUC & EER & TAR@1\% & TAR@5\% \\
\midrule
Naive VGG16~\cite{simonyan2014very}
& 59.61 & 42.94 & 1.83 & 7.45
& 53.76 & 46.92 & 1.28 & 5.98 \\
RST~\cite{fondje2020cross}
& 81.07 & 27.12 & 26.42 & 43.78
& 66.60 & 38.68 & 10.51 & 21.63 \\
DPIT-VGG16~\cite{fondje2022learning}
& 90.24 & 18.24 & 26.42 & 53.74
& 88.05 & 20.15 & 18.92 & 44.69 \\
Swin-S + Linear
& 96.41 & 9.48 & 37.36 & 76.02
& 92.16 & 15.99 & 26.63 & 57.75 \\
DPIT-ResNet50~\cite{fondje2022learning}
& 96.21 & 10.40 & 47.04 & 77.80
& \textbf{93.56} & \textbf{14.16} & \textbf{31.89} & \textbf{63.29} \\
ViT-B + Linear
& 97.51 & 7.65 & \textbf{49.81} & \textbf{86.73}
& 92.12 & 15.67 & 25.35 & 56.87 \\
DINOv2-B + Linear
& \textbf{97.52} & \textbf{7.63} & 49.03 & 86.37
& 92.29 & 15.62 & 27.26 & 57.61 \\
\bottomrule
\end{tabular}
\end{adjustbox}
\end{table*}

\begin{table*}[tbp]
\scriptsize
\centering
\caption{Cross-modal verification under protocol B at gallery scale, under frontal
(F$\rightarrow$F) and pose (F$\rightarrow$Pose) probes. Best results per column
are shown in \textbf{bold}. EER is lower-is-better ($\downarrow$); all others
higher-is-better ($\uparrow$).}
\label{tab:verification_protoX}
\setlength{\tabcolsep}{5pt}

\renewcommand{\arraystretch}{1.0}
\begin{adjustbox} {max width=0.8\textwidth}
\begin{tabular}{l cccc @{\hspace{2.5em}} cccc}
\toprule
& \multicolumn{4}{c}{F$\rightarrow$F (frontal)} & \multicolumn{4}{c}{F$\rightarrow$Pose (yaw)} \\
\cmidrule(lr){2-5}\cmidrule(lr){6-9}
Model & AUC & EER & TAR@1\% & TAR@5\%
      & AUC & EER & TAR@1\% & TAR@5\% \\
\midrule
Naive VGG16~\cite{simonyan2014very}
& 60.21 & 42.73 & 1.38 & 7.03
& 58.60 & 44.04 & 1.56 & 7.77 \\
RST~\cite{fondje2020cross}
& 84.08 & 24.39 & 30.78 & 48.04
& 72.04 & 34.57 & 13.44 & 26.64 \\
DPIT-VGG16~\cite{fondje2022learning}
& 91.56 & 16.62 & 29.79 & 57.74
& 88.78 & 19.55 & 18.58 & 45.85 \\
Swin-S + Linear
& 96.45 & 9.41 & 36.40 & 76.00
& 92.91 & 15.11 & 27.88 & 60.32 \\
DPIT-ResNet50~\cite{fondje2022learning}
& 96.78 & 9.50 & 50.46 & 80.71
& \textbf{94.68} & \textbf{12.73} & \textbf{36.08} & \textbf{67.53} \\
ViT-B + Linear
& \textbf{97.68} & \textbf{7.26} & \textbf{51.10} & \textbf{87.38}
& 93.44 & 14.12 & 28.97 & 62.45 \\
DINOv2-B + Linear
& 97.51 & 7.61 & 47.69 & 86.26
& 93.52 & 14.09 & 30.11 & 61.82 \\
\bottomrule
\end{tabular}
\end{adjustbox}
\end{table*}

\subsection{Thermal-Visible Face Verification}

Verification is evaluated using AUC, Equal Error Rate (EER), True Acceptance Rate (TAR) at False Acceptance Rate (FAR) of 1\% and 5\%, under Protocol ~A ($N=888$) and Protocol ~B($900$). Gallery scale directly affects low-FAR performance: larger galleries introduce more hard impostors near the decision threshold, making TAR@1\%FAR the primary metric for assessing deployment reliability. Tables~\ref{tab:verification_top7} 
and~\ref{tab:verification_protoX} report frontal and 
off-pose verification results for protocols A and B, respectively, while Figures~\ref{fig:roc_frontal} and ~\ref{fig:roc_yaw} show corresponding receiver operating characteristic (ROC) curves for the selected models under protocol~A.

Under frontal conditions (Table~\ref{tab:verification_top7}), 
transformer models achieve strong AUC (96.21--97.52\%), with 
DINOv2\,+\,Linear recording the best AUC (97.52\%), EER (7.63\%), 
 while ViT-Linear leads on TAR@1\%FAR (49.81\%) and TAR@5\%FAR (86.73\%).  RST records moderate performance over the naive baseline but remains below DPIT and the transformer models, while Naive VGG16 remains near random. Under {off-pose} conditions (Table~\ref{tab:verification_top7}), defined as any probe with yaw greater than $0\degree$ and up to $\pm 60\degree$, 
performance drops across all models, but the magnitude and pattern of degradation differ clearly across architectures. The ranking reverses: DPIT-ResNet50 leads all metrics (AUC 93.56\%, EER 14.16\%, TAR@1\%FAR 31.89\%, TAR@5\%FAR 63.29\%), while transformer models, which dominate under frontal conditions, suffer larger losses. RST also degrades noticeably but retains moderate performance, whereas the Naive VGG16 baseline collapses toward near-random behavior. 

This difference reflects how the models respond to pose variation. Transformer models, pretrained on large visible datasets, rely more on global facial structure and are more sensitive to geometric misalignment. Under pose changes, their embeddings shift toward the impostor region, and at $N = 888$ even small shifts are enough to reduce low-FAR performance. DPIT-ResNet50 relies on more localized features, making it less sensitive to pose changes and more stable across conditions. RST also degrades under pose variation, but retains moderate performance, indicating that its residual mapping captures some cross-spectral alignment, though it remains sensitive to geometric changes, while the Naive VGG16 baseline highlights the absence of meaningful cross-spectral representation.

Comparing the two protocols highlights the effect of enrollment diversity on pose
robustness. Protocol~A enrolls frontal images only, so an off-pose probe must
bridge both the modality gap and a pose gap absent from the gallery; Protocol~B
enrolls all capture conditions, so the averaged template already encodes
non-frontal appearance. Off-pose verification consequently improves for every
model: DPIT-ResNet50 gains in AUC (93.56\%$\rightarrow$94.68\%) and TAR@1\%FAR
(31.89\%$\rightarrow$36.08\%), and the transformers improve comparably (e.g.\
ViT-B TAR@1\%FAR 25.35\%$\rightarrow$28.97\%). As the protocols use different
subject sets ($N = 888$ vs.\ $N = 900$), these are trend comparisons rather than
exact effect sizes, but the direction is consistent: multi-perspective enrollment
narrows the frontal--pose gap that frontal-only enrollment leaves exposed.

These results highlight a key limitation of small-scale benchmarks: high AUC does not guarantee reliable low-FAR performance at scale. Transformer models perform well under frontal conditions, but DPIT shows greater robustness under pose variation, pointing to pose sensitivity as a critical challenge. Our dataset exposes this gap directly, demonstrating the importance of large-scale evaluation for uncovering failure modes hidden at smaller gallery sizes.

\begin{table*}[tbp]
\scriptsize  
\centering
\caption{Closed-set identification  under protocol A at two gallery scales, under
frontal (F$\rightarrow$F) and pose (F$\rightarrow$Pose) probes. Best result per
column in \textbf{bold}.}
\label{tab:identification_top7}
\setlength{\tabcolsep}{5pt}
\renewcommand{\arraystretch}{1.0}
\begin{adjustbox} {max width=0.8\textwidth}
\begin{tabular}{l cccccc @{\hspace{2.5em}} cccccc}
\toprule
& \multicolumn{6}{c}{F$\rightarrow$F (frontal)} & \multicolumn{6}{c}{F$\rightarrow$Pose (yaw)} \\
\cmidrule(lr){2-7}\cmidrule(lr){8-13}
& \multicolumn{3}{c}{$N{=}100$} & \multicolumn{3}{c}{$N{=}888$}
& \multicolumn{3}{c}{$N{=}100$} & \multicolumn{3}{c}{$N{=}888$} \\
\cmidrule(lr){2-4}\cmidrule(lr){5-7}\cmidrule(lr){8-10}\cmidrule(lr){11-13}
Model & R-1 & R-5 & R-10 & R-1 & R-5 & R-10
      & R-1 & R-5 & R-10 & R-1 & R-5 & R-10 \\
\midrule
Naive VGG16~\cite{simonyan2014very}
& 2.02 & 8.38 & 17.44 & 0.11 & 1.20 & 3.03
& 1.36 & 6.54 & 12.32 & 0.11 & 0.68 & 1.29 \\
RST~\cite{fondje2020cross}
& 32.74 & 60.80 & 71.50 & 5.33 & 34.38 & 41.80
& 12.66 & 31.14 & 42.48 & 2.62 & 12.34 & 16.72 \\
DPIT-VGG16~\cite{fondje2022learning}
& 28.98 & 61.36 & 76.00 & 10.64 & 26.71 & 36.29
& 20.28 & 51.82 & 68.10 & 7.00 & 17.96 & 26.21 \\
Swin-S + Linear
& 43.30 & 81.28 & 93.54 & 15.36 & 38.09 & 51.61
& 26.46 & 61.10 & 77.62 & 7.47 & 21.15 & 31.59 \\
DPIT-ResNet50~\cite{fondje2022learning}
& 55.90 & 86.34 & 93.46 & \textbf{27.42} & 52.14 & 64.19
& \textbf{38.14} & \textbf{71.78} & \textbf{83.78} & \textbf{15.80} & \textbf{34.89} & \textbf{45.83} \\
ViT-B + Linear
& 55.88 & 90.00 & 96.62 & 23.25 & 52.82 & 67.50
& 26.32 & 59.70 & 76.80 & 8.10 & 23.20 & 34.03 \\
DINOv2-B + Linear
& \textbf{58.20} & \textbf{92.20} & \textbf{97.50} & 23.85 & \textbf{57.96} & \textbf{72.68}
& 27.36 & 61.86 & 77.80 & 8.56 & 24.23 & 34.88 \\
\bottomrule
\end{tabular}
\end{adjustbox}
\end{table*}

\begin{table*}[tbp]
\scriptsize
\centering
\caption{Closed-set identification (Rank-$k$) under Protocol~B (Exhaustive) at
two gallery scales, under frontal (F$\rightarrow$F) and pose (F$\rightarrow$Pose)
probes. Best result per column in \textbf{bold}.}
\label{tab:identification_protoA}
\setlength{\tabcolsep}{5pt}
\renewcommand{\arraystretch}{1.0}
\begin{adjustbox} {max width=0.8\textwidth}
\begin{tabular}{l cccccc @{\hspace{2.5em}} cccccc}
\toprule
& \multicolumn{6}{c}{F$\rightarrow$F (frontal)} & \multicolumn{6}{c}{F$\rightarrow$Pose (yaw)} \\
\cmidrule(lr){2-7}\cmidrule(lr){8-13}
& \multicolumn{3}{c}{$N{=}100$} & \multicolumn{3}{c}{$N{=}900$}
& \multicolumn{3}{c}{$N{=}100$} & \multicolumn{3}{c}{$N{=}900$} \\
\cmidrule(lr){2-4}\cmidrule(lr){5-7}\cmidrule(lr){8-10}\cmidrule(lr){11-13}
Model & R-1 & R-5 & R-10 & R-1 & R-5 & R-10
      & R-1 & R-5 & R-10 & R-1 & R-5 & R-10 \\
\midrule
Naive VGG16~\cite{simonyan2014very}
& 2.62 & 10.18 & 17.40 & 0.48 & 1.07 & 2.78
& 2.00 & 8.78 & 16.14 & 0.24 & 0.88 & 2.05 \\
RST~\cite{fondje2020cross}
& 38.50 & 63.21 & 73.34 & 17.66 & 37.33 & 45.22
& 18.23 & 38.33 & 50.60 & 7.36 & 16.24 & 21.65 \\
DPIT-VGG16~\cite{fondje2022learning}
& 32.30 & 61.50 & 76.75 & 13.40 & 30.08 & 39.82
& 23.20 & 52.28 & 69.09 & 7.89 & 20.19 & 29.32 \\
Swin-S + Linear
& 37.61 & 80.15 & 92.63 & 13.49 & 35.23 & 47.54
& 26.16 & 61.32 & 78.95 & 7.56 & 21.96 & 31.45 \\
DPIT-ResNet50~\cite{fondje2022learning}
& 56.06 & 86.04 & 93.39 & \textbf{29.70} & 54.69 & 65.85
& \textbf{42.27} & \textbf{74.67} & \textbf{86.02} & \textbf{18.91} & \textbf{38.54} & \textbf{49.70} \\
ViT-B + Linear
& 52.19 & 88.01 & 95.13 & 21.12 & 51.21 & 65.65
& 30.15 & 64.90 & 79.29 & 9.21 & 26.08 & 37.67 \\
DINOv2-B + Linear
& \textbf{56.41} & \textbf{92.05} & \textbf{97.60} & 24.57 & \textbf{57.83} & \textbf{72.36}
& 31.24 & 69.09 & 83.16 & 10.43 & 28.02 & 39.94 \\
\bottomrule
\end{tabular}
\end{adjustbox}
\end{table*}

\subsection{Thermal-Visible Face Identification}

We also evaluated closed-set identification using VTF900, with visible gallery images and thermal probes, with each probe identity guaranteed to appear in the gallery. For each probe, cosine similarity is computed against the gallery template(s) and the results are reported using Rank-1, Rank-5, and Rank-10 accuracy across different gallery sizes under protocol A and protocol B, measuring how retrieval performance degrades as the identification problem scales. Tables~\ref{tab:identification_top7} 
and~\ref{tab:identification_protoA} report frontal and off-pose results for Protocols~A and B, respectively.

Increasing the gallery size reveals an insightful trend, as all models consistently suffer significant degradation. Under frontal conditions, Rank-1 accuracy drops by 26--34 percentage points between $N = 100$ and $N = 900$ across all evaluated models. This behavior is largely a structural effect of identification; however, the extent of this degradation depends strongly on the quality of the learned representations, as seen in the differing robustness across models. Models that appear robust at small scale become significantly less reliable at larger gallery scale. Under frontal conditions (Table~\ref{tab:identification_top7}), DINOv2\,+\, Linear records a Rank-1 accuracy of 58.20\% vs DPIT-ResNet50's 55.90\% at $N = 100$; as the gallery grows, DPIT-ResNet50 reverses this gap, reaching 27.42\% versus DINOv2's 23.85\% at
$N = 888$. This crossover reflects a fundamental change in what determines Rank-1 performance as gallery size grows. At small galleries, broadly discriminative embeddings are sufficient to rank the correct identity first. At large galleries, distinguishing among the most similar identities is a challenge, and that is where DPIT shows stronger robustness.  

The effect of pose further amplifies this behavior. Under 
off-pose conditions 
(Table~\ref{tab:identification_top7}), all models suffer additional degradation beyond the scaling effect, but the transformer models are affected disproportionately. At $N = 100$, the gap between DPIT-ResNet (38.14\%) and the best transformer (DINOvs, 27.36\%) is already 11 points, wider than any gap observed under frontal conditions at any scale. As the gallery grows, this gap does not close; at $N = 888$, DPIT-ResNet achieves 15.80\% while all transformer models fall below 10.16\%, with the weakest reaching only 7.47\%. Relative to chance ($\approx 0.11\%$ at $N = 888$), the weakest transformer
reaches roughly 66$\times$ chance and DPIT-ResNet50 about 140$\times$; Naive VGG16 stays closest to chance, and RST gives only modest gains, a meaningful practical spread. The complete gallery-scaling results across $N \in \{100, 200, 400, 888/900\}$ for both protocols and probe conditions are reported in the supplementary material section §A.3. 

Enrollment diversity again narrows the pose gap. Under Protocol~B's
multi-perspective enrollment, off-pose Rank-1 at $N = 100$ improves for every
learned model relative to Protocol~A---DPIT-ResNet50 $38.14\% \rightarrow
42.27\%$, DINOv2 $27.36\% \rightarrow 31.24\%$, ViT-B $26.32\% \rightarrow
30.15\%$ (Table~\ref{tab:identification_protoA})---consistent with the
verification trend, though again across different subject sets. Collectively,
scale transforms the difficulty of cross-spectral identification in ways small
galleries cannot reveal. The widening off-pose advantage of DPIT-ResNet50 across
scales, and its frontal crossover at large scale, indicate that robustness to
both modality shift and geometric misalignment grows more critical as galleries
scale---the conditions real deployment demands, and which, to our knowledge,
VTF900 is the first visible--thermal benchmark large enough to evaluate directly.

\section{Conclusion}
We introduced VTF900, a visible--thermal face dataset that is the largest by subject count (\textit{to the best of our knowledge}) and the first large enough to evaluate cross-spectral recognition under realistic gallery sizes and pose variation. Across verification and closed-set identification, performance degraded consistently as the gallery grows, and pose variation amplifies this effect, degradations that remained hidden at the small scales of prior benchmarks. Transformer models exceled under frontal probes but degrade rapidly with scale, particularly at the low-FAR operating points where genuine--impostor separation matters most, whereas DPIT-ResNet50 remained stable at large-scale and off-pose conditions;  and Naive VGG16 confirms the difficulty of matching. We also showed that multi-perspective enrollment narrows the frontal--pose gap, pointing to enrollment design as a practical lever for robustness. Overall, robustness to scale and pose---not only peak accuracy---was
the decisive factor towards operational cross-spectral recognition, and VTF900 provides the benchmark to measure it.






{\textbf{Acknowledgements:} This work was supported in part by Cyan Systems, Inc., under U.S. Department of Defense contract W909MY-19-C-0019. The views and conclusions expressed herein are those of the authors and should not be interpreted as necessarily representing the official policies, either expressed or implied, of Cyan Systems, Inc., or the U.S. Government. The authors also thank John Caulfield and Orrin Wellman for their support, discussion, and assistance regarding this work.}

{\small
\bibliographystyle{ieeenat_fullname}
\bibliography{main}
}

\clearpage
\appendix

\section{Supplementary Material}

This supplementary material provides dataset, training, and evaluation details
omitted from the main paper for space, along with additional experimental
results. It is organised as follows:

\begin{enumerate}
\item \textbf{Detailed VTF900 Description} (\S\ref{sec:supp:data}): per-condition image
      counts, paired cross-spectral subject coverage, and image counts for the
      two enrollment protocols.
\item \textbf{Training configuration} (\S\ref{sec:supp:train}): the full
      hyperparameter and augmentation specification.
\item \textbf{Experimental results} (\S\ref{sec:supp:results}): gallery-scaling
      results for both protocols, the multi-template ($K$) ablation, and the
      verification template-construction analysis for Protocol~B.
\end{enumerate}

All results use the same models and evaluation code as the main paper.

\subsection{Detailed VTF900 Description}
\label{sec:supp:data}

VTF900 comprises 900 subjects, 103,587 visible and 136,947 broadband (thermal)
images across eight capture conditions, grouped into \emph{Frontal}; \emph{Pose}
(Yaw-Left/Right, Pitch-Up/Down); \emph{Expression}; and \emph{Occlusion}
(Glasses, Mask). Occlusion is the union of glasses and mask, counting subjects in
both once.

Table~\ref{tab:supp:coverage} reports each group's \emph{paired} coverage of
subjects with an image of that condition in \emph{both} modalities, the only
population that can be matched cross-spectrally. Frontal and Pose are near-universal (900
and 895 paired subjects), so the scaling analysis rests on these full-coverage
conditions. Expression and Occlusion, captured opportunistically, reach only 247
(27.4\%) and 182 (20.2\%) paired subjects and are treated as supplementary
covariates.

\begin{table}[h]
\footnotesize
\centering
\caption{VTF900 condition groups: image counts and \emph{paired} subject coverage.
``Paired'' counts subjects with $\geq 1$ image in \emph{both} modalities.
Subject percentages are of 900; image percentages are of the modality total.}
\label{tab:supp:coverage}
\setlength{\tabcolsep}{4pt}
\renewcommand{\arraystretch}{1.1}
\begin{tabular}{l rr rr rrr}
\toprule
& \multicolumn{2}{c}{Visible imgs} & \multicolumn{2}{c}{Broadband imgs}
& \multicolumn{3}{c}{Subjects} \\
\cmidrule(lr){2-3}\cmidrule(lr){4-5}\cmidrule(lr){6-8}
Group & $n$ & \% & $n$ & \% & Vis & Bb & Paired (\%) \\
\midrule
Frontal    & 50,168 & 48.4 & 83,122 & 60.7 & 900 & 900 & 900 (100.0) \\
Pose       & 24,615 & 23.8 & 29,790 & 21.8 & 896 & 898 & 895 (99.4) \\
Expression & 14,651 & 14.1 &  4,596 &  3.4 & 728 & 305 & 247 (27.4) \\
Occlusion  & 14,153 & 13.7 & 19,439 & 14.2 & 192 & 219 & 182 (20.2) \\
\midrule
\textbf{Total} & \textbf{103,587} & \textbf{100.0} & \textbf{136,947} & \textbf{100.0}
& \textbf{900} & \textbf{900} & \textbf{900 (100.0)} \\
\bottomrule
\end{tabular}
\end{table}

Table~\ref{tab:supp:poseoccl} breaks these groups down. Yaw is near-universal
(891 paired subjects, 99.0\%), while pitch reaches only 188 (20.9\%). Within
Occlusion, glasses dominate (148 paired subjects) over masks (14), and glasses
subjects are far more densely sampled, so the two are not interchangeable.

\begin{table}[t]
\footnotesize
\centering
\caption{Pose and occlusion breakdown by subject count. ``Paired'' is defined as
in Table~\ref{tab:supp:coverage}; percentages are of the 900 subjects.}
\label{tab:supp:poseoccl}
\setlength{\tabcolsep}{5pt}
\renewcommand{\arraystretch}{1.1}
\begin{tabular}{l rr rr rr}
\toprule
& \multicolumn{2}{c}{Visible} & \multicolumn{2}{c}{Broadband} & \multicolumn{2}{c}{Paired} \\
\cmidrule(lr){2-3}\cmidrule(lr){4-5}\cmidrule(lr){6-7}
Condition & $n$ & \% & $n$ & \% & $n$ & \% \\
\midrule
\multicolumn{7}{l}{\textit{Pose}} \\
Yaw-Left            & 891 & 99.0 & 891 & 99.0 & 886 & 98.4 \\
Yaw-Right           & 884 & 98.2 & 888 & 98.7 & 879 & 97.7 \\
Yaw, either side    & 892 & 99.1 & 897 & 99.7 & 891 & 99.0 \\
Yaw, both sides     & 883 & 98.1 & 882 & 98.0 & 874 & 97.1 \\
Pitch-Up            & 164 & 18.2 & 126 & 14.0 & 105 & 11.7 \\
Pitch-Down          & 251 & 27.9 & 225 & 25.0 & 127 & 14.1 \\
Pitch, either       & 287 & 31.9 & 292 & 32.4 & 188 & 20.9 \\
Pitch, both         & 128 & 14.2 &  59 &  6.6 &  47 &  5.2 \\
Any pose            & 896 & 99.6 & 898 & 99.8 & 895 & 99.4 \\
\midrule
\multicolumn{7}{l}{\textit{Occlusion}} \\
Glasses only        & 160 & 17.8 & 168 & 18.7 & 148 & 16.4 \\
Mask only           &  21 &  2.3 &  32 &  3.6 &  14 &  1.6 \\
Both glasses \& mask&  11 &  1.2 &  19 &  2.1 &   7 &  0.8 \\
Any occlusion       & 192 & 21.3 & 219 & 24.3 & 182 & 20.2 \\
\bottomrule
\end{tabular}
\end{table}

\paragraph{Enrollment protocols.}
We release both enrollment protocols for reproducibility.

\textbf{Protocol A ($G_{10}$/$P_{10}$)} is the fixed-count protocol. 10 visible
frontal images are enrolled per subject, and each subject is probed with 20
broadband images: 10 frontal and 10 pose (Yaw-Left and Yaw-Right combined),
covering $N = 888$ subjects, 8,880 gallery and 17,760 probe images. Pose,
expression, and occlusion are excluded from the gallery, matching the ARL-VTF
visible-frontal baseline. Selection uses metadata only, no embeddings, model, or
quality score, so it is model-independent, and every model is scored on identical
images. The three sets are released as \texttt{gallery\_frontal.txt},
\texttt{probes\_frontal.txt}, and \texttt{probes\_pose.txt}, with identical
subject lists so the pose gap is directly interpretable.

\textbf{Protocol B ($G_{all}$/$P_{all}$)} is the all-conditions upper bound:
every visible image is enrolled, and every broadband image probed, giving
$N = 900$ subjects, 103,587 gallery and 136,947 probe images. It applies no
selection and measures what each system achieves given all enrollment data.
Gallery enrollment averages 115 images per subject and is deliberately
heterogeneous, so a subject template averages across all conditions. The two
sets are released as \texttt{all\_visible.txt} and \texttt{all\_broadband.txt}.

Two points bear on interpretation. First, 715 of the 900 subjects have a single
visit, so the 10 image Protocol~A enrollment draws 10 frames from one session
(sampled to span the capture) rather than ten independent enrollments; the benefit
of enrollment richness is therefore smaller, given the limited intra-subject
variation. Second, the protocols use different subject counts ($N = 888$ vs.\
$N = 900$), so their absolute numbers are not directly comparable, and we do not
compare them as such.

\paragraph{Provenance.}
Every manifest header records the source CSV, its modification time, and the
first sixteen hex digits of its SHA-256; every embedding cache stores the same
value and is refused at load if it disagrees. A protocol file therefore cannot be
silently paired with a metadata file that has changed.
\subsection{Training Configuration}
\label{sec:supp:train}

All transformer variants share an identical training protocol, so differences in the
benchmark tables are attributable to backbone and head choice alone. Models are
fine-tuned on ARL-VTF development split~1 (236 training, 59 validation identities)
and evaluated on VTF900; no VTF900 data is seen during training. Batches are drawn
by a \texttt{PairedIdentitySampler} (16 identities per batch, matched
visible--thermal pairs at 1:1), guaranteeing positive pairs so the cross-modal
contrastive term is well-posed. Table~\ref{tab:supp:hparams_compact} lists complete hyperparameter details.

\subsection{Experimental Results}
\label{sec:supp:results}

This section reports the full closed-set identification results across all gallery
scales for both protocols  (Tables~\ref{tab:idA_ff}--\ref{tab:idB_fp}), the
multi-template ($K$) ablation (Table~\ref{tab:kablation}), and the verification
template-construction analysis under Protocol~B
(Table~\ref{tab:verifstrategies}).

\subsubsection{Number of gallery templates ($K$)}
\label{sec:kablation}

Table~\ref{tab:kablation} reports identification with multi-template galleries:
each identity is represented by $K$ $k$-means centroids, and a probe is scored by
the maximum cosine similarity over them; $K{=}1$ is the single mean-pooled
template used throughout. Larger $K$ gives no consistent gain---improvements are
small and model/probe-dependent, while cost grows linearly with $K$. On the
strongest backbones (DINOv2-B, DPIT-ResNet50), $K{=}1$ is best or tied at the
all-probe and frontal conditions; the one exception is DINOv2-B under pose, where
$K{=}4$ adds 2.6 Rank-1 points. We adopt $K{=}1$ throughout.

\subsubsection{Verification template construction}
\label{sec:verifstrategies}

Table~\ref{tab:verifstrategies} ablates verification-template construction,
comparing three ways of forming verification trials under Protocol~B for the
selected backbones with an image-level gallery. \emph{Image-to-image} scores
every probe against every gallery image ($1.42{\times}10^{10}$ pairs);
\emph{image-to-template} scores each probe against a subject template with 500
sampled impostors per probe; \emph{template-to-template} scores one probe
template against one gallery template per subject. Gallery-side aggregation helps
consistently: image-to-template improves every metric over image-to-image for all
five models. Probe-side aggregation is conditional---template-to-template improves
every metric for the four strongest systems (lifting ViT-B TAR@1\%FAR from
47.30\% to 59.44\%) but regresses on DPIT-VGG16's frontal AUC and TAR, where its
per-image embeddings are less consistent. Pooling helps only when the pooled
embeddings are already consistent. Because genuine-trial counts differ by four
orders of magnitude, TAR@FAR is not comparable across strategies; we report
image-to-template in the main text as the balance between the two. In
Table~\ref{tab:verifstrategies}, \emph{All} denotes every probe condition,
\emph{Fr.}\ only frontal broadband probes, and \emph{Po.}\ only yaw (left and
right) probes.


\begin{table*}[t]
\small
\centering
\caption{Detailed Training specification }
\label{tab:supp:hparams_compact}
\renewcommand{\arraystretch}{1.25}
\begin{tabularx}{\textwidth}{@{}l@{\hspace{14em}} >{\raggedright\arraybackslash}X@{}}
\toprule
\multicolumn{2}{@{}l}{\textit{Architecture}} \\
\cmidrule(r){1-2}
Framework      & Dual-stream, weight-shared backbone \\
Heads          & Modality-specific projection heads \\
Backbones      & ViT-B/16, Swin-S, DINOv2 ViT-B \\
Initialisation & ImageNet-sup.\ (ViT-B, Swin-S); self-sup.\ (DINOv2) \\
Embedding      & 512-d, $\ell_2$-normalised \\
Input          & $224\times224$, dropout 0.3 \\
Loss heads     & linear, ArcFace, CosFace, AdaFace ($s{=}64$, $m{=}0.5$) \\
Classes        & 236 (ARL-VTF development split 1) \\
\midrule
\multicolumn{2}{@{}l}{\textit{Optimisation}} \\
\cmidrule(r){1-2}
Optimiser      & AdamW, weight decay 0.05 \\
Learning rate  & $5\times10^{-4}$ (heads), $2.5\times10^{-5}$ (backbone) \\
Schedule       & linear warmup 5 epochs $\rightarrow$ cosine decay \\
Batch          & $16\times2$ accumulation $=32$ effective \\
Identities     & 16 per batch (paired sampler) \\
Regularisation & grad-clip 1.0, label smoothing 0.1 \\
Training       & $\leq$60 epochs, early-stop patience 7 (val.\ Rank-1) \\
Convergence    & epochs 20--43 \\
Hardware       & $4\times$ RTX~2080, \texttt{DataParallel} \\
\midrule
\multicolumn{2}{@{}l}{\textit{Objective} \; $\mathcal{L}=\lambda_{ce}\mathcal{L}_{ce}+\lambda_{cm}\mathcal{L}_{cm}$} \\
\cmidrule(r){1-2}
$\mathcal{L}_{ce}$ & classification, both streams; $\lambda_{ce}=0.5$ \\
$\mathcal{L}_{cm}$ & cross-modal InfoNCE; $\lambda_{cm}=1.0$, $\tau=0.05$ \\
Domain adapt.  & disabled for every reported run \\
\midrule
\multicolumn{2}{@{}l}{\textit{Augmentation} (identical on both streams)} \\
\cmidrule(r){1-2}
Resize         & $256\times256$ \\
Crop           & \texttt{RandomCrop(224)} \\
Flip           & \texttt{RandomHorizontalFlip(0.5)} \\
Channels       & grayscale replicated to 3 channels \\
Normalise      & ImageNet statistics \\
Erasing        & \texttt{RandomErasing(0.25)} \\
Evaluation     & \texttt{Resize(224)}, 3-channel, \texttt{Normalize}; no aug. \\
\bottomrule
\end{tabularx}
\end{table*}

\clearpage


\begin{table*}[t]
\footnotesize
\centering
\caption{Closed-set identification under protocol A at four gallery scales, under Frontal $\rightarrow$ Frontal. Best per column in \textbf{bold}}
\label{tab:idA_ff}
\setlength{\tabcolsep}{4pt}
\renewcommand{\arraystretch}{1.1}
\begin{adjustbox}{max width=\textwidth}
\begin{tabular}{l ccc @{\hspace{1.4em}} ccc @{\hspace{1.4em}} ccc @{\hspace{1.4em}} ccc}
\toprule
& \multicolumn{3}{c}{$N{=}100$} & \multicolumn{3}{c}{$N{=}200$} & \multicolumn{3}{c}{$N{=}400$} & \multicolumn{3}{c}{$N{=}888$} \\
\cmidrule(lr){2-4}\cmidrule(lr){5-7}\cmidrule(lr){8-10}\cmidrule(lr){11-13}
Model & R-1 & R-5 & R-10 & R-1 & R-5 & R-10 & R-1 & R-5 & R-10 & R-1 & R-5 & R-10 \\
\midrule
Naive VGG16~\cite{simonyan2014very} & 2.02 & 8.38 & 17.44 & 1.02 & 5.09 & 8.71 & 0.26 & 3.13 & 5.44 & 0.11 & 1.20 & 3.03 \\
RST~\cite{fondje2020cross} & 32.74 & 60.80 & 71.50 & 21.79 & 49.08 & 59.22 & 11.55 & 41.58 & 50.30 & 5.33 & 34.38 & 41.80 \\
DPIT-VGG16~\cite{fondje2022learning} & 28.98 & 61.36 & 76.00 & 21.51 & 47.06 & 61.14 & 15.88 & 36.74 & 49.25 & 10.64 & 26.71 & 36.29 \\
Swin-S + Linear & 43.30 & 81.28 & 93.54 & 33.84 & 69.77 & 82.31 & 23.44 & 53.24 & 68.98 & 15.36 & 38.09 & 51.61 \\
DPIT-ResNet50~\cite{fondje2022learning} & 55.90 & 86.34 & 93.46 & 46.10 & 76.81 & 86.42 & \textbf{37.10} & 65.36 & 77.21 & \textbf{27.42} & 52.14 & 64.19 \\
ViT-B + Linear & 55.88 & 90.00 & 96.62 & 44.80 & 80.76 & 91.37 & 33.89 & 67.98 & 81.77 & 23.25 & 52.82 & 67.50 \\
DINOv2-B + Linear & \textbf{58.20} & \textbf{92.20} & \textbf{97.50} & \textbf{49.53} & \textbf{85.63} & \textbf{94.68} & 36.79 & \textbf{73.25} & \textbf{86.09} & 23.85 & \textbf{57.96} & \textbf{72.68} \\
\bottomrule
\end{tabular}
\end{adjustbox}
\end{table*}

\begin{table*}[t]
\footnotesize
\centering
\caption{Closed-set identification under protocol A at four gallery scales, under frontal $\rightarrow$ pose. Best per column in \textbf{bold}}
\label{tab:idA_fp}
\setlength{\tabcolsep}{4pt}
\renewcommand{\arraystretch}{1.1}
\begin{adjustbox}{max width=\textwidth}
\begin{tabular}{l ccc @{\hspace{1.4em}} ccc @{\hspace{1.4em}} ccc @{\hspace{1.4em}} ccc}
\toprule
& \multicolumn{3}{c}{$N{=}100$} & \multicolumn{3}{c}{$N{=}200$} & \multicolumn{3}{c}{$N{=}400$} & \multicolumn{3}{c}{$N{=}888$} \\
\cmidrule(lr){2-4}\cmidrule(lr){5-7}\cmidrule(lr){8-10}\cmidrule(lr){11-13}
Model & R-1 & R-5 & R-10 & R-1 & R-5 & R-10 & R-1 & R-5 & R-10 & R-1 & R-5 & R-10 \\
\midrule
Naive VGG16~\cite{simonyan2014very} & 1.36 & 6.54 & 12.32 & 0.53 & 3.26 & 6.35 & 0.24 & 1.46 & 3.19 & 0.11 & 0.68 & 1.29 \\
RST~\cite{fondje2020cross} & 12.66 & 31.14 & 42.48 & 8.96 & 23.79 & 32.16 & 4.64 & 16.83 & 23.74 & 2.62 & 12.34 & 16.72 \\
DPIT-VGG16~\cite{fondje2022learning} & 20.28 & 51.82 & 68.10 & 14.79 & 38.50 & 53.55 & 11.16 & 27.57 & 40.34 & 7.00 & 17.96 & 26.21 \\
Swin-S + Linear & 26.46 & 61.10 & 77.62 & 18.98 & 45.72 & 62.56 & 11.93 & 33.23 & 47.25 & 7.47 & 21.15 & 31.59 \\
DPIT-ResNet50~\cite{fondje2022learning} & \textbf{38.14} & \textbf{71.78} & \textbf{83.78} & \textbf{30.73} & \textbf{59.54} & \textbf{72.09} & \textbf{23.21} & \textbf{47.59} & \textbf{59.87} & \textbf{15.80} & \textbf{34.89} & \textbf{45.83} \\
ViT-B + Linear & 26.32 & 59.70 & 76.80 & 19.99 & 49.19 & 64.02 & 13.01 & 35.74 & 49.30 & 8.10 & 23.20 & 34.03 \\
DINOv2-B + Linear & 27.36 & 61.86 & 77.80 & 21.11 & 50.80 & 66.37 & 14.23 & 36.31 & 50.71 & 8.56 & 24.23 & 34.88 \\
\bottomrule
\end{tabular}
\end{adjustbox}
\end{table*}


\begin{table*}[t]
\footnotesize
\centering
\caption{Closed-set identification under protocol B at four gallery scales, under frontal $\rightarrow$ frontal. Best per column in \textbf{bold}}
\label{tab:idB_ff}
\setlength{\tabcolsep}{4pt}
\renewcommand{\arraystretch}{1.1}
\begin{adjustbox}{max width=\textwidth}
\begin{tabular}{l ccc @{\hspace{1.4em}} ccc @{\hspace{1.4em}} ccc @{\hspace{1.4em}} ccc}
\toprule
& \multicolumn{3}{c}{$N{=}100$} & \multicolumn{3}{c}{$N{=}200$} & \multicolumn{3}{c}{$N{=}400$} & \multicolumn{3}{c}{$N{=}900$} \\
\cmidrule(lr){2-4}\cmidrule(lr){5-7}\cmidrule(lr){8-10}\cmidrule(lr){11-13}
Model & R-1 & R-5 & R-10 & R-1 & R-5 & R-10 & R-1 & R-5 & R-10 & R-1 & R-5 & R-10 \\
\midrule
Naive VGG16~\cite{simonyan2014very} & 2.62 & 10.18 & 17.40 & 1.65 & 5.64 & 10.09 & 0.77 & 3.05 & 5.81 & 0.48 & 1.07 & 2.78 \\
RST~\cite{fondje2020cross} & 38.50 & 63.21 & 73.34 & 31.72 & 56.38 & 65.63 & 22.79 & 45.86 & 55.60 & 17.66 & 37.33 & 45.22 \\
DPIT-VGG16~\cite{fondje2022learning} & 32.30 & 61.50 & 76.75 & 24.05 & 51.94 & 66.20 & 18.71 & 39.93 & 52.41 & 13.40 & 30.08 & 39.82 \\
Swin-S + Linear & 37.61 & 80.15 & 92.63 & 29.14 & 65.02 & 80.71 & 22.24 & 49.91 & 65.14 & 13.49 & 35.23 & 47.54 \\
DPIT-ResNet50~\cite{fondje2022learning} & 56.06 & 86.04 & 93.39 & 48.32 & 78.30 & 87.27 & \textbf{39.86} & 67.38 & 78.55 & \textbf{29.70} & 54.69 & 65.85 \\
ViT-B + Linear & 52.19 & 88.01 & 95.13 & 41.62 & 79.28 & 89.41 & 32.36 & 68.08 & 80.76 & 21.12 & 51.21 & 65.65 \\
DINOv2-B + Linear & \textbf{56.41} & \textbf{92.05} & \textbf{97.60} & \textbf{49.01} & \textbf{85.57} & \textbf{94.43} & 37.62 & \textbf{73.67} & \textbf{85.79} & 24.57 & \textbf{57.83} & \textbf{72.36} \\
\bottomrule
\end{tabular}
\end{adjustbox}
\end{table*}

\begin{table*}[t]
\footnotesize
\centering
\caption{Closed-set identification under protocol B at four gallery scales, under frontal $\rightarrow$ pose. Best per column in \textbf{bold}}
\label{tab:idB_fp}
\setlength{\tabcolsep}{4pt}
\renewcommand{\arraystretch}{1.1}
\begin{adjustbox}{max width=\textwidth}
\begin{tabular}{l ccc @{\hspace{1.4em}} ccc @{\hspace{1.4em}} ccc @{\hspace{1.4em}} ccc}
\toprule
& \multicolumn{3}{c}{$N{=}100$} & \multicolumn{3}{c}{$N{=}200$} & \multicolumn{3}{c}{$N{=}400$} & \multicolumn{3}{c}{$N{=}900$} \\
\cmidrule(lr){2-4}\cmidrule(lr){5-7}\cmidrule(lr){8-10}\cmidrule(lr){11-13}
Model & R-1 & R-5 & R-10 & R-1 & R-5 & R-10 & R-1 & R-5 & R-10 & R-1 & R-5 & R-10 \\
\midrule
Naive VGG16~\cite{simonyan2014very} & 2.00 & 8.78 & 16.14 & 0.69 & 4.65 & 8.41 & 0.32 & 2.24 & 4.66 & 0.24 & 0.88 & 2.05 \\
RST~\cite{fondje2020cross} & 18.23 & 38.33 & 50.60 & 15.10 & 30.71 & 39.92 & 10.18 & 22.52 & 30.28 & 7.36 & 16.24 & 21.65 \\
DPIT-VGG16~\cite{fondje2022learning} & 23.20 & 52.28 & 69.09 & 17.20 & 43.10 & 57.92 & 11.93 & 29.96 & 42.67 & 7.89 & 20.19 & 29.32 \\
Swin-S + Linear & 26.16 & 61.32 & 78.95 & 19.32 & 47.03 & 63.34 & 12.00 & 33.11 & 47.46 & 7.56 & 21.96 & 31.45 \\
DPIT-ResNet50~\cite{fondje2022learning} & \textbf{42.27} & \textbf{74.67} & \textbf{86.02} & \textbf{33.00} & \textbf{62.74} & \textbf{75.90} & \textbf{26.09} & \textbf{50.80} & \textbf{63.39} & \textbf{18.91} & \textbf{38.54} & \textbf{49.70} \\
ViT-B + Linear & 30.15 & 64.90 & 79.29 & 21.06 & 51.68 & 66.75 & 14.31 & 38.63 & 52.17 & 9.21 & 26.08 & 37.67 \\
DINOv2-B + Linear & 31.24 & 69.09 & 83.16 & 24.51 & 55.98 & 70.93 & 16.45 & 41.18 & 55.68 & 10.43 & 28.02 & 39.94 \\
\bottomrule
\end{tabular}
\end{adjustbox}
\end{table*}

\begin{table*}[t]
\footnotesize
\centering
\caption{Effect of the number of per-identity gallery templates $K$ on Protocol~B
($G_{all}$/$P_{all}$) identification (Rank-$k$, \%). Each identity is represented by $K$
$k$-means centroids scored by max cosine similarity; $K{=}1$ is the single
mean-pooled template used throughout. Best per metric within each (model, probe)
block in \textbf{bold}}
\label{tab:kablation}
\setlength{\tabcolsep}{5pt}
\renewcommand{\arraystretch}{1.1}
\begin{tabular}{l c ccc @{\hspace{1.6em}} ccc @{\hspace{1.6em}} ccc}
\toprule
& & \multicolumn{3}{c}{All probes} & \multicolumn{3}{c}{Frontal} & \multicolumn{3}{c}{Pose} \\
\cmidrule(lr){3-5}\cmidrule(lr){6-8}\cmidrule(lr){9-11}
Model & $K$ & R-1 & R-5 & R-10 & R-1 & R-5 & R-10 & R-1 & R-5 & R-10 \\
\midrule
DPIT-VGG16~\cite{fondje2022learning} & 1 & \textbf{11.69} & \textbf{26.70} & \textbf{35.84} & 13.40 & \textbf{30.08} & \textbf{39.82} & 7.89 & \textbf{20.19} & \textbf{29.32} \\
& 2 & 11.41 & 26.22 & 34.56 & 13.11 & 29.18 & 38.12 & 7.68 & 19.59 & 28.00 \\
& 3 & 11.45 & 26.49 & 35.03 & 13.36 & 29.15 & 38.16 & 7.52 & 19.58 & 27.96 \\
& 4 & 11.57 & 26.43 & 34.92 & \textbf{13.58} & 29.36 & 38.31 & \textbf{7.94} & 19.32 & 27.33 \\
\midrule
Swin-S + Linear & 1 & 13.70 & 33.86 & 45.36 & 13.49 & 35.23 & 47.54 & \textbf{7.56} & \textbf{21.96} & 31.45 \\
& 2 & \textbf{14.36} & \textbf{34.17} & \textbf{46.03} & \textbf{14.23} & \textbf{35.40} & \textbf{48.01} & 7.26 & 21.55 & \textbf{31.72} \\
& 3 & 14.00 & 33.60 & 45.77 & 13.89 & 34.78 & 47.63 & 7.18 & 20.98 & 31.08 \\
& 4 & 13.72 & 33.48 & 45.90 & 13.53 & 34.40 & 47.67 & 7.23 & 21.23 & 31.25 \\
\midrule
ViT-B + Linear & 1 & 19.76 & 46.25 & 59.41 & \textbf{21.12} & \textbf{51.21} & \textbf{65.65} & 9.21 & 26.08 & 37.67 \\
& 2 & 20.49 & 46.53 & 59.92 & 21.06 & 49.87 & 64.65 & 9.74 & 26.61 & 38.30 \\
& 3 & \textbf{20.74} & 46.74 & 60.01 & 21.09 & 49.36 & 64.25 & \textbf{9.91} & \textbf{27.04} & 38.42 \\
& 4 & 20.72 & \textbf{47.32} & \textbf{60.77} & 20.91 & 49.86 & 64.74 & 9.60 & 27.02 & \textbf{38.73} \\
\midrule
DINOv2-B + Linear & 1 & \textbf{22.81} & \textbf{51.82} & 65.02 & \textbf{24.57} & \textbf{57.83} & \textbf{72.36} & 10.43 & 28.02 & 39.94 \\
& 2 & 22.26 & 51.14 & 64.68 & 22.72 & 55.09 & 69.54 & 12.26 & 32.33 & 45.07 \\
& 3 & 21.19 & 50.28 & 64.75 & 21.39 & 53.25 & 68.63 & 13.02 & 33.56 & 46.63 \\
& 4 & 21.86 & 50.84 & \textbf{65.29} & 22.36 & 53.50 & 69.10 & \textbf{13.04} & \textbf{34.23} & \textbf{47.26} \\
\midrule
DPIT-ResNet50~\cite{fondje2022learning} & 1 & \textbf{24.73} & \textbf{47.75} & \textbf{58.99} & \textbf{29.70} & \textbf{54.69} & \textbf{65.85} & \textbf{18.91} & \textbf{38.54} & 49.70 \\
& 2 & 23.60 & 46.88 & 58.50 & 28.38 & 53.58 & 65.29 & 18.03 & 38.36 & \textbf{49.91} \\
& 3 & 23.56 & 46.71 & 58.52 & 28.50 & 53.29 & 65.05 & 17.23 & 37.51 & 48.89 \\
& 4 & 23.80 & 46.93 & 58.31 & 28.58 & 53.59 & 64.94 & 17.17 & 36.91 & 48.21 \\
\bottomrule
\end{tabular}
\end{table*}

\begin{table*}[t]
\footnotesize
\centering
\caption{Protocol~B ($G_{all}$/$P_{all}$) verification under three scoring strategies
($N = 900$) Probe counts: All $n = 136{,}947$, Frontal
$n = 83{,}122$, Pose $n = 27{,}324$. Best strategy per metric (per model) in \textbf{bold};
EER lower is better.}
\label{tab:verifstrategies}
\setlength{\tabcolsep}{4pt}
\renewcommand{\arraystretch}{1.15}
\begin{adjustbox}{max width=\textwidth}
\begin{tabular}{l l ccc @{\hspace{1em}} ccc @{\hspace{1em}} ccc @{\hspace{1em}} ccc}
\toprule
& & \multicolumn{3}{c}{AUC $\uparrow$} & \multicolumn{3}{c}{EER $\downarrow$}
& \multicolumn{3}{c}{TAR@1\%FAR $\uparrow$} & \multicolumn{3}{c}{TAR@5\%FAR $\uparrow$} \\
\cmidrule(lr){3-5}\cmidrule(lr){6-8}\cmidrule(lr){9-11}\cmidrule(lr){12-14}
Model & Strategy & All & Fr. & Po. & All & Fr. & Po. & All & Fr. & Po. & All & Fr. & Po. \\
\midrule
DPIT-VGG16~\cite{fondje2022learning}
 & Image-to-image      & .8269 & .8820 & .8525 & 24.75 & 20.20 & 22.84 & 19.97 & 25.39 & 16.82 & 42.23 & 50.40 & 40.35 \\
 & Image-to-template   & .8709 & \textbf{.9156} & .8878 & 20.61 & 16.62 & 19.55 & 23.66 & \textbf{29.79} & 18.58 & 49.44 & \textbf{57.74} & 45.85 \\
 & Template-to-template & \textbf{.9222} & .9141 & \textbf{.9192} & \textbf{15.56} & \textbf{16.57} & \textbf{16.25} & \textbf{30.78} & 27.56 & \textbf{25.42} & \textbf{59.78} & 56.89 & \textbf{56.74} \\
\midrule
Swin-S + Linear
 & Image-to-image      & .9354 & .9454 & .9133 & 13.73 & 12.25 & 16.93 & 31.59 & 33.22 & 24.89 & 64.76 & 68.46 & 54.91 \\
 & Image-to-template   & .9539 & .9645 & .9291 & 11.11 & 9.41 & 15.11 & 35.40 & 36.40 & 27.88 & 72.08 & 76.00 & 60.32 \\
 & Template-to-template & \textbf{.9722} & \textbf{.9693} & \textbf{.9531} & \textbf{7.99} & \textbf{8.57} & \textbf{11.47} & \textbf{42.89} & \textbf{38.67} & \textbf{32.55} & \textbf{81.67} & \textbf{80.22} & \textbf{69.01} \\
\midrule
ViT-B + Linear
 & Image-to-image      & .9392 & .9520 & .9145 & 13.39 & 11.55 & 16.34 & 38.24 & 41.99 & 24.39 & 69.49 & 75.27 & 55.48 \\
 & Image-to-template   & .9632 & .9768 & .9344 & 9.93 & 7.26 & 14.12 & 47.30 & 51.10 & 28.97 & 80.38 & 87.38 & 62.45 \\
 & Template-to-template & \textbf{.9832} & \textbf{.9815} & \textbf{.9597} & \textbf{6.11} & \textbf{6.36} & \textbf{10.47} & \textbf{59.44} & \textbf{55.00} & \textbf{35.67} & \textbf{91.44} & \textbf{90.56} & \textbf{73.36} \\
\midrule
DINOv2-B + Linear
 & Image-to-image      & .9386 & .9523 & .9113 & 13.38 & 11.30 & 16.89 & 38.20 & 41.10 & 25.49 & 69.93 & 75.53 & 54.89 \\
 & Image-to-template   & .9615 & .9751 & .9352 & 10.19 & 7.61 & 14.09 & 44.55 & 47.69 & 30.11 & 78.90 & 86.26 & 61.82 \\
 & Template-to-template & \textbf{.9834} & \textbf{.9812} & \textbf{.9619} & \textbf{5.89} & \textbf{6.32} & \textbf{9.57} & \textbf{55.22} & \textbf{51.67} & \textbf{34.11} & \textbf{91.67} & \textbf{90.11} & \textbf{74.14} \\
\midrule
DPIT-ResNet50~\cite{fondje2022learning}
 & Image-to-image      & .9228 & .9410 & .9206 & 15.44 & 13.13 & 15.86 & 36.64 & 42.81 & 30.52 & 64.13 & 70.85 & 59.54 \\
 & Image-to-template   & .9539 & .9678 & .9468 & 11.61 & 9.50 & 12.73 & 44.30 & 50.46 & 36.08 & 74.11 & 80.71 & 67.53 \\
 & Template-to-template & \textbf{.9725} & \textbf{.9700} & \textbf{.9635} & \textbf{8.65} & \textbf{9.20} & \textbf{10.37} & \textbf{53.44} & \textbf{51.78} & \textbf{44.48} & \textbf{81.89} & \textbf{81.89} & \textbf{75.14} \\
\bottomrule
\end{tabular}
\end{adjustbox}
\end{table*}

\begingroup
\bibliographystyle{ieeenat_fullname}
\setbox0=\vbox{\small\bibliography{main}}
\endgroup


\end{document}